%% file: main.tex
\documentclass[10pt, a4paper]{article}

\usepackage[]{lrec-coling2024} % this is the new style

\usepackage{microtype}
\usepackage{graphicx}
\usepackage{xspace}
\usepackage{booktabs}
\usepackage{multirow}
\usepackage{amsmath}
\usepackage{sepfootnotes}
\usepackage{tablefootnote}
\usepackage{threeparttable}
\usepackage{enumitem}
\usepackage{arydshln}
\usepackage{color,soul}
\usepackage{CJKutf8}
\usepackage{hyperref}
\usepackage{natbib}

\newcommand{\dataset}{CroCoSum\xspace}

\newcommand{\symfootnote}[1]{%
\let\oldthefootnote=\thefootnote%
\stepcounter{mpfootnote}%
\addtocounter{footnote}{-1}%
\renewcommand{\thefootnote}{\fnsymbol{mpfootnote}}%
\footnote{#1}%
\let\thefootnote=\oldthefootnote%
}

\title{\dataset: A Benchmark Dataset for \\
Cross-Lingual Code-Switched Summarization}

\name{Ruochen Zhang$^{1}$\quad \quad Carsten Eickhoff$^{1,2}$\quad} 

\address{$^1$Brown University \\ $^2$University of Tübingen\\
ruochen\_zhang@brown.edu \quad c.eickhoff@acm.org \\}

\abstract{
Cross-lingual summarization (CLS) has attracted increasing interest in recent years due to the availability of large-scale web-mined datasets and the advancements of multilingual language models. However, given the rareness of naturally occurring CLS resources, the majority of datasets are forced to rely on translation which can contain overly literal artifacts. This restricts our ability to observe naturally occurring CLS pairs that capture organic diction, including instances of code-switching. This alteration between languages in mid-message is a common phenomenon in multilingual settings yet has been largely overlooked in cross-lingual contexts due to data scarcity. To address this gap, we introduce \dataset, a dataset of cross-lingual code-switched summarization of technology news. It consists of over 24,000 English source articles and 18,000 human-written Chinese news summaries, with more than 92\% of the summaries containing code-switched phrases. For reference, we evaluate the performance of existing approaches including pipeline, end-to-end, and zero-shot methods. We show that leveraging existing CLS resources as a pretraining step does not improve performance on \dataset, indicating the limited generalizability of current datasets. Finally, we discuss the challenges of evaluating cross-lingual summarizers on code-switched generation through qualitative error analyses.
 \\ \newline \Keywords{Code-Switching, Cross-Lingual, Summarization, Corpus, Collection, Dataset} }

\begin{document}
\begin{CJK*}{UTF8}{gbsn}
\maketitleabstract

\section{Introduction}

Cross-lingual summarization (CLS) is the task of producing summaries in a target language given source documents in a different language. CLS can help with the rapid dissemination of information across multiple languages in an increasingly globalized context. It is considered more challenging than within-language summarization, as it combines translation and summarization objectives~\citep{10.1162/tacl_a_00520}. With more multilingual resources~\citep{2019t5, laurenccon2022bigscience, scialom2020mlsum, hasan2021xl} becoming available and the advancement of large multilingual language models~\citep{liu2020multilingual, lin2021few, scao2022bloom}, CLS has attracted significant attention in recent years. One key factor that has been limiting the development of CLS is data scarcity. Therefore, current CLS resources~\citeplanguageresource{wang2022clidsum, zheng2022long, ladhak-etal-2020-wikilingua, hasan2021crosssum, clads-emnlp} heavily rely on automatic or manual translation rather than collecting texts organically written in cross-lingual fashion. However, translated texts have been reported to exhibit features different from the original language's composition~\citep{graham-etal-2020-statistical}. Summaries generated by models trained on such texts may contain instances of ``Translationese'', such as literal translations of idioms~\citep{wang2022understanding}. Humans, on the other hand, code-switch between languages, especially when there is no appropriate translation, or when readers are more familiar with the original foreign entity names or expressions. There have been summarization resources addressing the code-switching phenomenon~\citep{mehnaz2021gupshup}, but they focus on summarizing from already code-switched source texts.
\begin{figure*}[ht!]
\centering
    \includegraphics[width=0.95\linewidth]{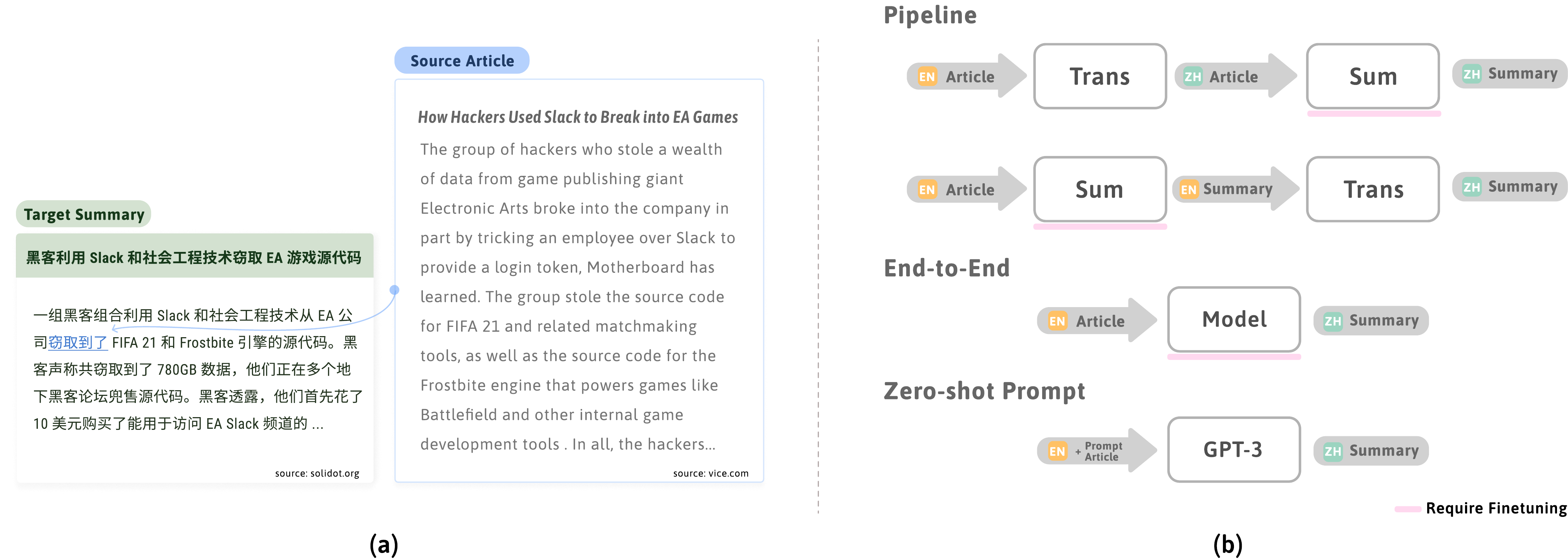}
    \caption{\label{fig:data_figure} (a) A Data Example of Source Article and Target Summary Pair. (b) Baseline Approaches.}
    \vspace{-2mm}
\end{figure*}

To study the phenomenon of code-switching in CLS, we introduce \dataset, a new benchmark dataset for \textbf{Cro}ss-Lingual \textbf{Co}de-Switched \textbf{Sum}marization containing human-written Chinese-English code-switched summaries of technology-focused news articles in English\footnote{Our collection and code can be accessed at \url{https://github.com/RosenZhang/CroCoSum}}. The code-switched summaries are gathered from \url{solidot.org}, an online platform for sharing technology-related news. The summaries are written and posted by real users including technical professionals, open-source enthusiasts and university students. They are then reviewed by the website's editors before being published. Each post contains one or more links pointing to the original news sources. We collect the original news articles from the Internet Archive\footnote{\url{https://archive.org/}} and only consider those sources as written in English. We then construct the source-target pairs by tracing back to posts referring to those English source articles. Our final dataset contains over 24,000 English source articles and over 18,000 code-switched summaries. More than 92\% of the summaries contain code-switched phrases and over 55\% of sentences within the summaries contain code-switching spans. A data example is shown in Figure~\ref{fig:data_figure}(a). 

We follow existing CLS approaches~\citeplanguageresource{zhu-etal-2019-ncls,ladhak-etal-2020-wikilingua, wang2022clidsum}, and evaluate baselines including pipeline, end-to-end and zero-shot processing on \dataset (See Figure~\ref{fig:data_figure}(b)). Pipeline methods can be further broken down into translate-then-summarize and summarize-then-translate approaches. We use the Google Translate API as the translation module in the pipeline methods. For the summarizer module and the end-to-end method, we experiment with various pretrained multilingual sequence-to-sequence models such as mT5~\citep{xue2020mt5}, mBART~\citep{liu2020multilingual} and mBART-50~\citep{tang2020multilingual}. We also prompt GPT-3~\citep{brown2020language} to generate summaries in a zero-shot manner. Among our baselines, end-to-end finetuning mBART-50 yields the best results. However, we notice a decrease in performance when leveraging other CLS resources as a pretraining step in our best baseline, indicating limited generalizability provided by current CLS resources. Finally, by comparing various automatic metrics, we observe no clear relationship between summarization quality and code-switching complexity, calling for future research designing a more comprehensive evaluation framework. 

The novel contributions of this work are three-fold: 1) we introduce \dataset, the first collection designed for examining the phenomenon of code-switching in cross-lingual summarization, 2) we provide an initial set of benchmark performance measurements of various baseline approaches and architectures (pipeline, end-to-end and zero-shot prompting), 3) we perform a qualitative analysis revealing the common error types in code-switched generation and highlighting opportunities for future investigation.

\section{Related Work}
\paragraph{Cross-Lingual Summarization} Early cross-lingual summarization resources like En2ZhSum and Zh2EnSum~\citeplanguageresource{zhu-etal-2019-ncls} have been constructed via machine translation from originally monolingual summarization datasets. Collections such as ClidSum~\citep{wang2022clidsum} crowdsource human translations to obtain cross-lingual resources of higher quality. Additionally, with the prevalence of large-scale web-mined texts, some works~\citeplanguageresource{ladhak-etal-2020-wikilingua, clads-emnlp, hasan2021crosssum} focus on the number of languages covered and exploit websites that provide multilingual content, such as WikiHow and Wikipedia. For CLS approaches, due to the limited availability of parallel corpora, early works~\citep{wan-etal-2010-cross, 7502066, 8370729, zhu-etal-2019-ncls} use pipeline methods to break CLS into two subtasks: translation and summarization then develop dedicated models for each subtask. With the availability of large-scale parallel corpora and pretrained multilingual language models, more works~\citep{ladhak-etal-2020-wikilingua, hasan2021crosssum, wang2022clidsum} experiment with end-to-end approaches with different pretraining techniques, allowing models to directly take source texts in one language and summarize them in another. However, all current CLS resources, whether translated or web-mined,  may contain different levels of ``Translationese'' artifacts due to some dependency on machine-translated texts~\citep{wang2022understanding,ladhak-etal-2020-wikilingua}. \citet{10.1162/tacl_a_00447} mention that automatically crawled and filtered datasets tend to show a lower quality compared to hand-produced collections. Furthermore, unlike human-written summaries crafted for each article, using initial sentences or paragraphs as summaries in one language does not ensure alignment with the source texts sampled from articles in another language. Despite many efforts in resource collection, none of the existing collections acknowledge the phenomenon of code-switching, alternating languages in mid-message, in cross-lingual summaries. \dataset, therefore, is different from existing resources in that all summaries are written and reviewed by humans to meet publishing standards. It provides an ideal testbed in which to observe organic human diction in CLS settings.
\paragraph{Code-switching} Since code-switching is observed more frequently in colloquial (rather than formal) texts~\citep{dogruoz-etal-2021-survey, winata2022decades},  it is challenging to gather large-scale well-annotated resources to study this phenomenon~\citep{yong2023prompting}. Established code-switching resources are usually collected from social media texts and focus on sequence tagging applications, including, for example, language identification~\citep{das-gamback-2014-identifying, barman-etal-2014-code}, NER~\citep{singh-etal-2018-named}, and POS-tagging~\citeplanguageresource{aguilar-etal-2020-lince}. There have also been works that develop objective metrics to describe the level of code-switching complexity~\citep{gamback2014measuring, gamback-das-2016-comparing, khanuja-etal-2020-gluecos}. Besides sequence tagging, other works also touch on short-form generation~\citep{Mondal_Pathak_Jyothi_Raghuveer} and speech recognition tasks with audio data~\citeplanguageresource{li2012mandarin, li2022talcs, lovenia-etal-2022-ascend}. Gupshup~\citeplanguageresource{mehnaz2021gupshup}, to the best of our knowledge, is the only collection dedicated to studying code-switching in summarization~\citep{dougruoz2023survey}. Different from \dataset, it introduces code-switched source texts by translating the SAMSum~\citeplanguageresource{gliwa-etal-2019-samsum} dataset into Hinglish instead of studying this phenomenon in organically occurring target summaries. 
\input{cls_stats.tex}
\input{csmetrics_stats.tex}

\section{\dataset}
\dataset contains 18,557 human-written Chinese-English code-switched summaries and 24,171 English source articles. More than 92\% of the summaries, and 55\% of sentences in the summaries contain code-switched phrases. In the sections below, we describe our data collection and comparison with existing resources in detail.

\subsection{Dataset Construction}
\label{sec:construction}
The target summaries in \dataset are collected from \url{solidot.org}, an Chinese online platform for IT professionals and open-source enthusiasts to share technology-related news. Users summarize technology news from international outlets and compose short Chinese summary posts. Due to the highly timely nature of news and tech-focused topics, some English entities and phrases in the original news items are yet to receive formal translations or are preferred in their original form by website writers and readers. To guarantee a high-quality feed, prior to getting published on the platform, each post is encouraged to contain at least one hyperlink pointing to the original news source for credibility (See~\ref{fig:data_figure} for an example.) and is reviewed by human editors to ensure clarity. Our initial crawl contains 28,953 post webpages and 51,258 embedded links. We obtain web pages of the embedded links from the Internet Archive and use the newspaper3k\footnote{\url{https://github.com/codelucas/newspaper}} package to extract titles and articles. After running language detection\footnote{\url{https://github.com/Mimino666/langdetect}} on the extracted content, we only retain English sources (80\% of all sources). From the remaining websites, we filter out those that contain failed extractions (empty body, login information, javascript and cookie notifications, etc.) based on a manually curated list of cue words and sentences. We then trace back and remove posts containing these deleted websites and posts that do not contain any links, resulting in the final collection of 18,557 posts and 24,171 corresponding English source articles. We construct source-target pairs by matching summary posts with the embedded links to the source articles. If an article contains multiple sources, a list with all source texts is mapped to that article. 
See Appendix~\ref{sec:examples} for data examples. 

To examine the data quality obtained from the automatic filtering process, we recruit 3 college graduates with bilingual proficiency and ask them to annotate a random sample of 20 source-target pairs following \citetlanguageresource{clads-emnlp}. Here, we depart from the original authors' annotation scheme of seeking binary answers to two general questions, and instead collect more fine-grained ratings of both the syntactic and semantic dimensions of the data instances. We adopt 4 rubrics suggested by \citetlanguageresource{grusky2018newsroom}, which are Fluency(F), Coherence(C), Informativeness(I) and Relevance(R). The pairs were rated using a 5-point Likert scale with 1 being the lowest score. In our annotations, the scores are F: 4.62, C:4.95, I: 3.97 and R: 4.18. Notably, 75\% of our samples received a rating of 4 and 5 for the semantic dimensions (I and R), aligning with the quality statistics observed in \citetlanguageresource{clads-emnlp}. Finally, the collected data is partitioned into distinct training (70\%), validation (15\%) and test (15\%) sets.

\subsection{Dataset Characterization}
\paragraph{CLS Dataset Statistics}
\label{sec:stats}
Table \ref{tab:cls_stats} shows a comparison of key measurements between \dataset and other existing CLS resources\footnote{XSAMSum and XMediaSum40k are subsets of ClidSum~\citeplanguageresource{wang2022clidsum} with CLS data. The remaining MediaSum424k subset only contains monolingual data.} containing English-Chinese source-target pairs. Note that, except En2ZhSum and \dataset, all other datasets contain multiple languages among either their sources or targets, but the statistics are only calculated on their English-Chinese subset for a more accurate comparison. Additionally, since \dataset could contain more than one source article per summary, we concatenate multiple sources for such examples before calculation. Besides general dataset descriptions like construction types, domains, dataset sizes and languages, we provide the average number of words and sentences (segmented by stanza\footnote{\url{https://github.com/stanfordnlp/stanza}}) in English source texts and Chinese target summaries. We observe that, compared with CLS datasets in the news domain (on average 600-700+ words per sample), \dataset contains much more expansive source texts (1,000+ words per sample). Summaries in \dataset are also much longer than those in other CLS datasets (225.6 words vs.\ 20-80 words on average per summary). 

\paragraph{Code-Switching Complexity}
We also investigate the code-switching frequency of \dataset in comparison with other code-switching datasets. Because there is no existing code-switched resource for CLS, we extend our comparison to the loosely related summarization dataset GupShup~\citep{mehnaz2021gupshup} that focuses on summarizing from Hindi-English code-switched source texts. For a more comprehensive analysis, we also select datasets that contain Chinese-English code-switched texts but across different tasks such as language identification (LID) in tweets~\citep{solorio-etal-2014-overview} and speech recognition~\citeplanguageresource{lyu2010seame, lovenia-etal-2022-ascend}. Although current CLS resources do not study the code-switching phenomenon, they could potentially contain code-switched tokens in their target summaries. Therefore, we include them in our comparison as well. We use the metrics suggested by~\citet{gamback-das-2016-comparing} below to measure the level of code-switching complexity. 

\textit{Code-Mixing Index (CMI)} is the fraction of language-dependent tokens not belonging to the matrix language (the most frequent language in the sentence) in the utterance. CMI for a sentence $x$ can be computed as 
\begin{flalign*}
&\textit{CMI}(x) = \begin{cases}
     \frac{N(x) - \max_{L_i \in L}{\{t_{L_i}\}(x)}}{N(x)} & : N(x) > 0\\
     0 & : N(x) = 0
\end{cases} &
\end{flalign*}

where $N(x)$ refers to the number of tokens in sentence $x$ except the language-independent \footnote{Tokens shared by languages. For example, numerical digits.}tokens, and $t_{L_i}$ refer to tokens in language $L_i$. For monolingual sentences, CMI is 0. Higher CMIs indicate more code-switched tokens. 

\textit{Intra-Sentence Switch Points (SP)} are the number of word boundaries within a sentence for which the words on either side are in different languages. 

Both metrics measure sentence-level switching. We report dataset-level statistics in Table \ref{tab:csmetrics_stats} by taking an average of sentences across all examples (All) and in those with code-switched words (Switched). Additionally, we provide the total number of sentences (Total Sents), average sentence length (Avg Sent Len), the number of code-switched sentences (Switched-Sents) and their percentage (Switched-\%). Notice that among all summarization datasets, cross-lingual ones report metrics based on their target summaries, while GupShup bases them on its source documents.

We observe that \dataset offers the highest percentage of code-switched sentences, overall CMI, and SP among all CLS datasets. To our surprise, CLS datasets such as WikiLingua contain high CMI for their code-switched summaries. We hypothesize that this is due to their shorter summary length which makes code-switched tokens relatively more prominent. However, its scores in other code-switching metrics are significantly lower compared to \dataset. 

When comparing to code-switching datasets, we note that in terms of CMI over code-switched examples, \dataset shows a lower score compared to speech corpora like ASCEND and SEAME. We assume that this stems from ASCEND/SEAME's sentences of colloquial text being on average shorter and of less formal diction than what is observed in the curated news domain. Yet our dataset still has a similar or higher percentage of code-switched sentences and SP.

To summarize, \dataset is the only CLS dataset that studies code-switching in human-written target summaries. It has longer source and target texts compared to other CLS datasets in the news domain, and a similar level of code-switching complexity compared to existing code-switching datasets in other tasks.

\section{Experiments and Evaluations}
\input{baseline.tex}

In this section, we describe the details of our baseline approaches and metrics used for evaluation.
\subsection{Baselines}
Similar to previous works~\citeplanguageresource{nguyen-daume-iii-2019-global, ladhak-etal-2020-wikilingua, wang2022clidsum}, our baselines include pipeline methods that decompose the CLS task into machine translation and summarization as well as end-to-end approaches with direct cross-lingual supervision. Given the recent convincing performance of prompt learning with large language models, we also experiment with zero-shot prompting of GPT-3 to understand the dataset's difficulty and the necessity of training dedicated models. 

\paragraph{Pipeline}
Pipeline methods decompose the CLS task into summarization and machine translation subtasks. The main reason behind employing such two-step processes in earlier works was a lack of cross-lingual resources at the time~\citep{ladhak-etal-2020-wikilingua}. Depending on the ordering of subtasks, methods can be further broken down into translate-then-summarize and summarize-then-translate approaches. We choose Google's translation API as \citet{wang2022clidsum} find it to perform best in pipeline methods. We use it via the Translators\footnote{\url{https://github.com/uliontse/translators}} library as our translation module. In the summarization module, we finetune the same multilingually pretrained models described below in the end-to-end baseline.
    \begin{itemize}[nolistsep]
        \item Translate-then-Summarize (Trans-Sum). We first translate all English source texts into Chinese. Then we finetune different models on the summarization task with the translated source texts and original Chinese code-switched summaries in our training set.
        \item Summarize-then-Translate (Sum-Trans). We finetune multilingual models with English source texts and English summaries translated from the Chinese ground truths. At inference time, the generated English summary is translated back into Chinese for final evaluation.
    \end{itemize}
    
\paragraph{End-to-End} The end-to-end method requires models to learn translation and summarization at the same time in a supervised manner. More specifically, the model takes in articles in the source language (English) and is expected to generate a summary in the target language (Chinese) directly. We adopt the following multilingually pretrained models as our cross-lingual summarizers.
    \begin{itemize}[nolistsep]
        \item mT5~\citep{xue2020mt5} is a multilingual variant of T5~\citep{raffel2020exploring} that was pretrained in 101 languages unsupervisedly. We use mT5-base with 580M parameters to be closer in size to the other models below.
        \item mBART~\citep{liu2020multilingual} is a sequence-to-sequence model using denoising objectives for neural machine translation. We use mBART25 which was trained on a 25-language monolingual corpus, and contains 610M parameters. 
        \item mBART-50~\citep{tang2020multilingual} is an extension to mBART, adding tokens for additional languages in its embedding layer, and pretraining on a total of 50 languages. It is of the same size as mBART25. 
    \end{itemize}

\paragraph{Zero-shot Prompting} Different from the methods above, the zero-shot method requires no training and relies on the models' generalizability to unseen tasks through manually crafted prompts. In this approach, we format each source-target pair in the test set as ``Can you summarize the English article below in Chinese? <English source text>'' and feed it into GPT-3~\citep{brown2020language} and GPT-3.5. It is expected to generate Chinese summaries following the prompts without task-specific training. 

In the pipeline and end-to-end baselines, using the train set of \dataset, we finetune the summarization models based on their implementations in the transformers library~\citep{wolf-etal-2020-transformers} for 15 epochs on a single RTX 3090 GPU and select the best checkpoint for final evaluation. 
\footnote{See Appendix~\ref{sec:exp_details} for more experiment details.}
\subsection{Evaluation Metrics}
In Table \ref{tab:baseline}, we report F1 scores of ROUGE-\{1,2, L\} (R1/R2/RL)~\citep{lin-2004-rouge} and BERTScore (BS)~\citep{zhang2019bertscore} to compare the lexical and semantic similarities between the predicted and ground truth summaries. We also compute basic statistics such as the average number of sentences (Sents.) and word counts (Words). Differences in code-switched sentence percentage ($\text{Sents}_{cs}$ \%) and code-switching metrics ($\text{CMI}_{all/cs}$, $\text{SP}_{all/cs}$) with respect to gold summaries in the test set are also reported for more comprehensive analysis.
\input{augmentation_results.tex}

\section{Results and Analysis}
\paragraph{Automatic Metrics} Among all baseline methods, end-to-end finetuning generally attains the best performance in terms of both ROUGE and BERTScore. mBART50, specifically, works best compared to all other base models. Our results seem to contradict what \citet{ladhak-etal-2020-wikilingua} and \citet{wang2022clidsum} have reported, namely pipeline methods perform better than end-to-end finetuning methods. However, note that their dataset is able to supply gold monolingual article-summary pairs for training the summarizers by exploiting web-mined parallel resources or human translations while ours relies on silver pairs from the Google Translation API. This makes our pipeline methods more prone to error propagation, and therefore may lead to worse performance. This discrepancy in results suggests that end-to-end methods could provide promising results when there are no monolingual resources for training individual submodules.  

We find zero-shot prompting with GPT-3's performance comparable to pipeline approaches with finetuned mT5 and GPT-3.5 comparable to that with finetuned mBART. This result is encouraging as GPT-3 is only exposed to a small number of non-English tokens mixed in its English pretraining data, and it shows that simple prompts can elicit useful knowledge for unseen tasks without any specific training. However, the proprietariness of the GPT-3.5 model obstructs a better understanding of its improvements over its predecessor. 

For different multilingually pretrained base models, we find that mBART50 and mBART have similar performance across different baselines and consistently outperform mT5. We hypothesize that this distinction may stem from the fact that mBART and mBART50 have already been pretrained for translation tasks, whereas mT5 has only been pretrained unsupervisedly, requiring additional effort to establish alignments between languages. 
\begin{figure}
    \centering
    \resizebox{\columnwidth}{!}{%
    \includegraphics{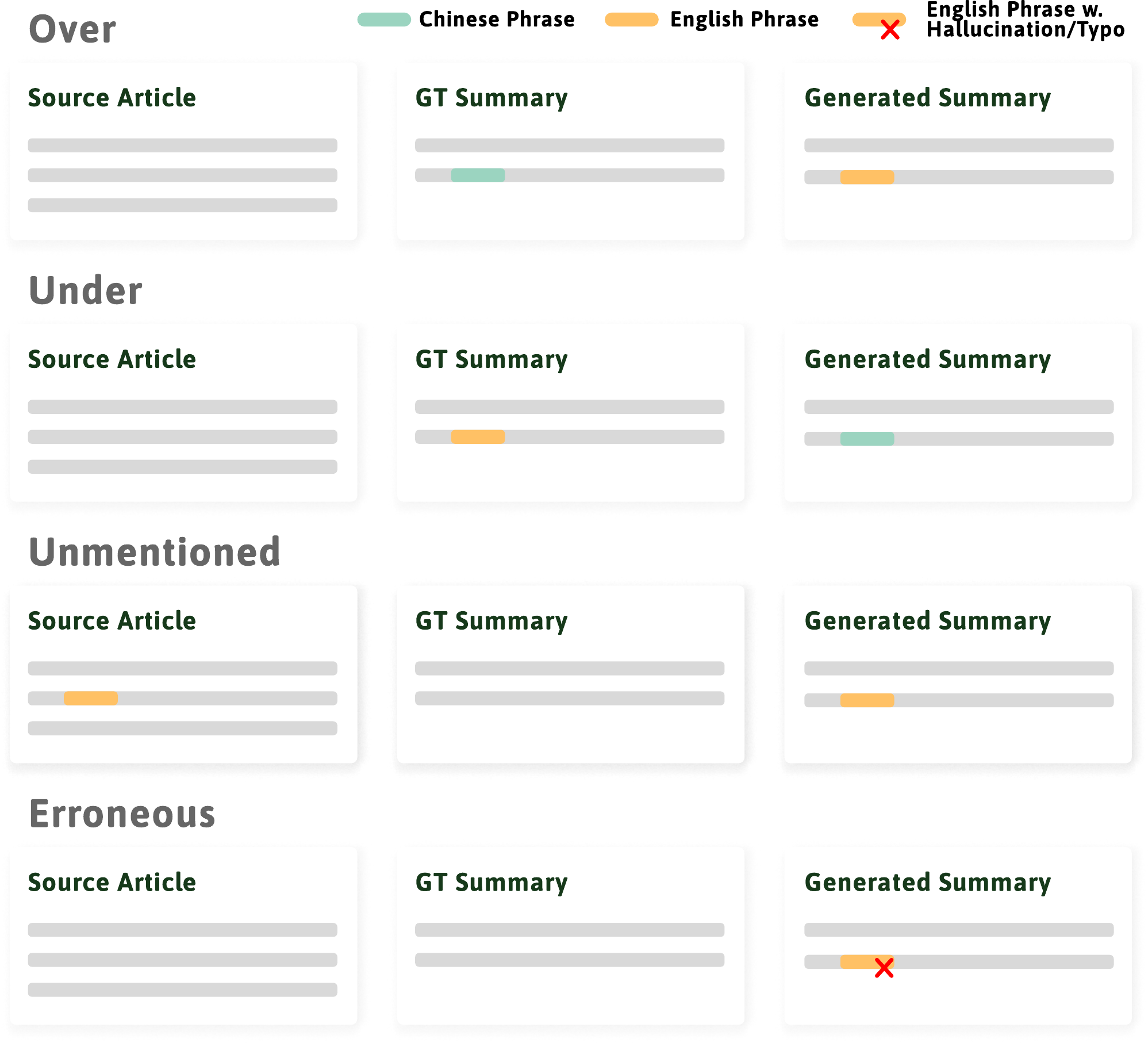}%
    }
    \caption{Illustration of Error Types.}
    \label{fig:error_types}%
\end{figure}

\paragraph{Data Augmentation} To explore whether we can further boost our best-performing baseline, we leverage English-Chinese CLS pairs in WikiLingua and CrossSum in an additional pretraining step for mBART50 before finetuning it on \dataset. We only select the web-mined datasets instead of those created by translation as they contain more natural texts, and minimize the occurrences of translationese in the datasets. We experiment with three settings: 1) pretrain on WikiLingua or CrossSum then finetune on \dataset 2) pretrain on the shuffled set of WikiLingua and CrossSum, then finetune on \dataset 3) finetune on the shuffled set of WikiLingua, CrossSum and \dataset. Results in Table \ref{tab:augmentation} show that pretraining with additional CLS pairs does not improve the model's performance on \dataset, but rather results in a slight score decrease, indicating limited generalizability provided by existing CLS datasets and motivating the need for more diverse CLS resources.

\paragraph{Code-switching Metrics}
Besides automatic metrics of summarization quality, we also compute code-switching metrics for the ground truth summaries and model generations using code-switched percentage, CMI and SP in Table \ref{tab:baseline}. The smaller the difference in metrics, the more closely we assume the model prediction to resemble how humans choose to code-switch in their writing. Yet, from the results, we note that the smallest differences are obtained by different baseline approaches as well as different pretrained base models. While the end-to-end finetuned mBART50 excels in automatic metrics, its level of code-switching is not the closest to the ground truth in any given code-switching metric. This discrepancy calls for more in-depth error analysis as current ngram-based auto metrics, limited to monolingual texts, fail to identify semantically correct instances like over/under-switched cases as described in Section~\ref{sec:quality-analysis} below.

\section{Qualitative Analysis}
\label{sec:quality-analysis}
To further investigate the challenges in generating code-switched summaries under CLS settings, we randomly select 100 test generations provided by our best-performing baseline and manually compare them to the corresponding ground truth summaries. We propose four error types to categorize the differences in code-switching tokens between model predictions and human-written summaries. See Figure~\ref{fig:error_types} for a graphical representation of the error types. Each prediction may contain zero, one or more of the errors below.

\paragraph{Over Switched Phrases} In 30 out of the 100 examples, the generated summaries contain English phrases that should have been Chinese according to the ground truth. We find that the generations tend to follow the three patterns below:
\begin{figure}[h!]
    \centering
    \resizebox{\columnwidth}{!}{%
    \includegraphics{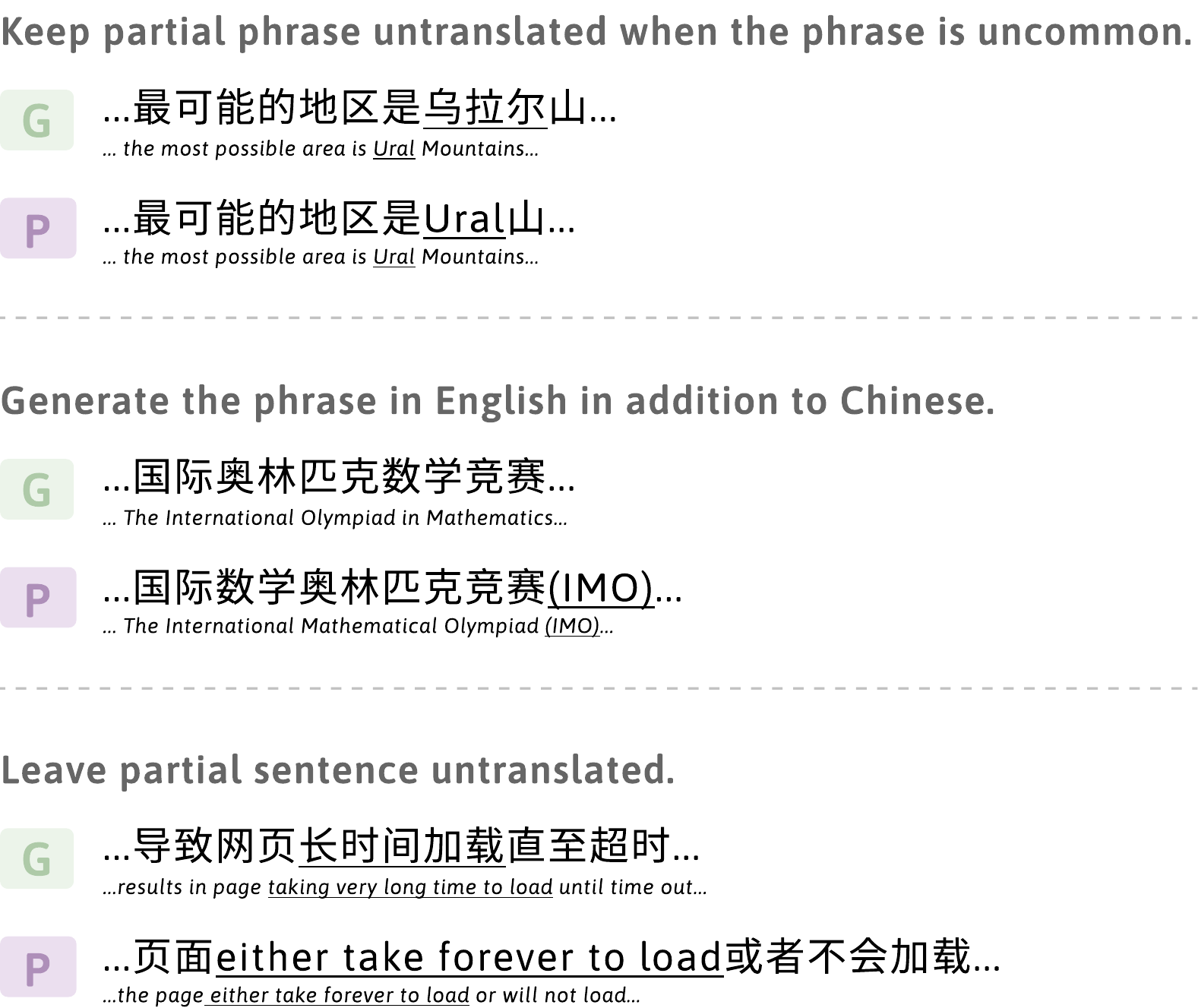}%
    }
\end{figure}
% \begin{enumerate}
%     \item Keep partial phrase untranslated when the phrase is uncommon.\\
%     G: ...最可能的地区是\underline{乌拉尔山}... \\
%     P: ...有可能的地区是\underline{Ural 山}...
%     \item Generate the phrase in the source language in addition to that in the target language. \\
%     G: ...国际奥林匹克数学竞赛...\\
%     P: ...国际数学奥林匹克竞赛\underline{(IMO)}...
%     \item Leave partial sentence untranslated. \\
%     G: ...导致网页\underline{长时间加载}直至超时... \\
%     P: ...页面\underline{either take forever to load}或者不会加载... 
% \end{enumerate}
\paragraph{Under Switched Phrases} 8 of the 100 generated summaries contain phrases in the target language which the ground truths chose to code-switch into the source language.
\begin{figure}[h!]
    \centering
    \resizebox{\columnwidth}{!}{%
    \includegraphics{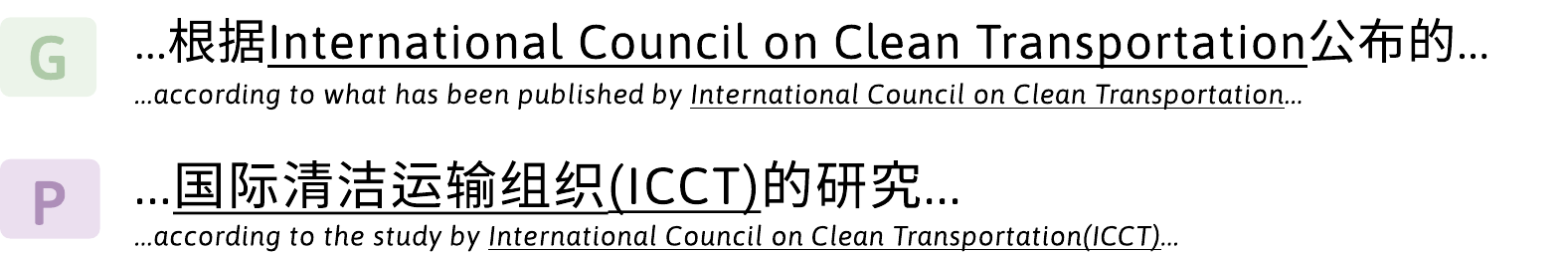}%
    }
\end{figure}
\paragraph{Unmentioned Code-switched Phrases} 47 summaries contained English phrases that exist in the source texts but not in the ground truth summaries. As the inputs are lengthy, the ground truths and predictions may focus on different aspects during summarization. In the example below, the prediction details the information source whereas the ground truth omits this information.

\begin{figure}[h!]
    \centering
    \resizebox{\columnwidth}{!}{%
    \includegraphics{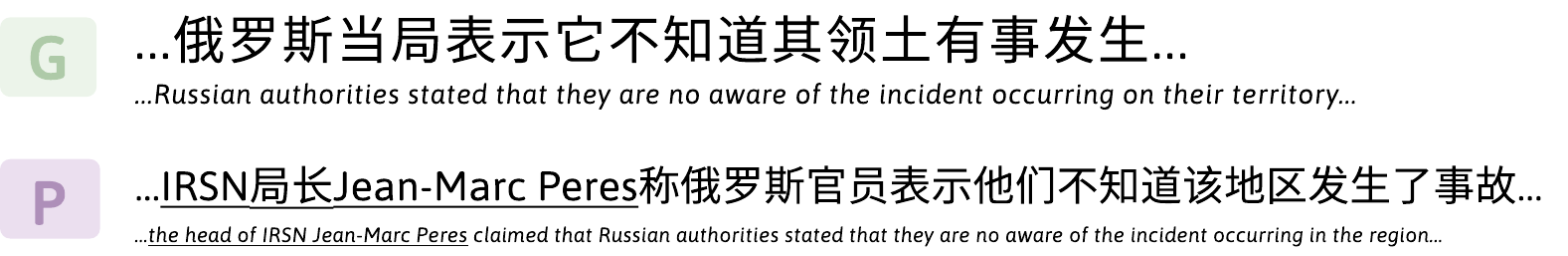}%
    }
\end{figure}
\paragraph{Erroneous Code-switched Phrases} In 43 examples, the model generates misspelled English phrases or those that contradict, or are irrelevant to the source text. For example, it wrongly attributes World Wide Web inventor Tim Berners-Lee, to be the CTO instead of Chris Urmson, who actually holds that position.
\begin{figure}[h!]
    \centering
    \resizebox{\columnwidth}{!}{%
    \includegraphics{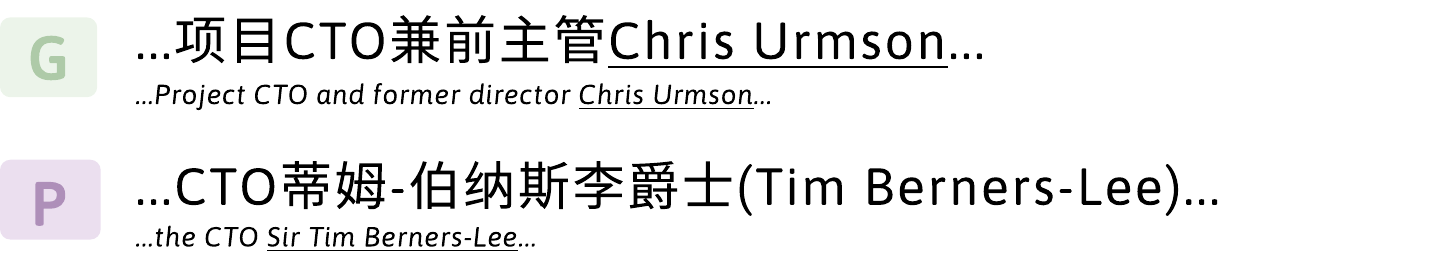}%
    }
\end{figure}

As an additional point of reference, we collect human assessments of Fluency, Coherence, Informativeness and Relevance for 20 summary generations to the same sets of articles sampled and described in Sec~\ref{sec:construction}: F 3.32, C 3.77, I 3.40, R 3.23. Compared to human reference summaries, these scores from model-generated summaries are notably lower, indicating a lesser quality than those produced by humans.

As shown in our qualitative analysis, \dataset reveals various complications in evaluating code-switched summary generation. Especially for the first three cases, when models produce factual statements either in languages different from or omitted by the ground truth, a more comprehensive semantic evaluation that involves human judgment on naturalness and relevance is required. 

\section{Conclusion}
In this paper, we introduce \dataset, the first collection of organic cross-lingual code-switched text summarization. \dataset distinguishes itself by exhibiting a significantly higher code-switching frequency when compared to existing CLS datasets, while still demonstrating comparable complexity to other non-summarization code-switching datasets. We provide benchmark performances of current CLS baseline approaches and an in-depth analysis highlighting the challenges of evaluating code-switched summaries using existing metrics. 

\section*{Limitations}
In our baseline experiments, observations are based on the model sizes allowed by our local compute resources. A more exhaustive analysis can be obtained by experimenting with greater baseline variation, including different model sizes, prompt templates, and few-shot experiments given a more generous compute budget. 

Additionally, the scope of code-switching presented in our paper is restricted to a range of topics covered by the data source. The predominant presentation of code-switching occurs in the form of named entities, such as scientific terminologies and product names. Using code-switched terms, rather than their official Chinese equivalents, is a common practice among the authors and the intended audience. This writing style is favored because it facilitates rapid dissemination of news delivery and promotes more straightforward understanding. This focus on named entity mentions, however does not trivialize the problem. The decision between translation and literal copying of phrases and names follows intricate linguistic and cultural rules and is far from arbitrary. 

Finally, given the diverse cultural contexts in which code-switching may occur, it is important to acknowledge that \dataset, and outputs from models finetuned on \dataset, may not fully encapsulate the complexities of actual code-switching practices among different linguistic and cultural backgrounds.

\section*{Acknowledgement}
This research is supported in part by ODNI and IARPA via the BETTER program (2019-
19051600004). The views and conclusions contained herein are those of the authors and should not be interpreted as necessarily representing the official policies, either expressed or implied, of ODNI, IARPA, or the U.S. Government. We would like to thank Ellie Pavlick, Arman Cohan, Shannon Shen, Ziwei Chen, Junyi Xiong, Qiulin Su and Health NLP Lab members at Brown for their constructive feedback and helpful discussion.
\nocite{*}
\section{Bibliographical References}\label{sec:reference}

\bibliographystyle{lrec-coling2024-natbib}
\bibliography{lrec-coling2024-example}

\section{Language Resource References}
\label{lr:ref}
\bibliographystylelanguageresource{lrec-coling2024-natbib}
\bibliographylanguageresource{languageresource}

\include{appendix}

\end{CJK*}
\end{document}

%% file: cls_stats.tex

\begin{table*}[ht]
\centering
\resizebox{1\linewidth}{!}{%
\begin{tabular}{@{}rrrrrrrrrr@{}}
\toprule
                             &                  &                 &               & \multicolumn{3}{c}{\textbf{Source}} & \multicolumn{3}{c}{\textbf{Target}} \\ \cmidrule(l){5-7} \cmidrule(l){8-10} 
\textbf{Type}                         & \textbf{Dataset} & \textbf{Domain} & \textbf{Size} & Lang.       & Words      & Sents     & Lang.       & Words      & Sents     \\ \midrule
\multirow{3}{*}{Translation} & \textbf{En2ZhSum}         & News            & 370,687       & En         & 755.0      & 40.6        & Zh         & 84.4       & 3.6         \\
                             & \textbf{XSAMSum}          & Dialogue        & 16,369        & En         & 97.7       & 12.1      & De/Zh      & 32.0       & 2.0       \\
                             & \textbf{XMediaSum40k}     & Dialogue        & 40,000        & En         & 1661.5     & 113.5     & De/Zh      & 28.04      & 1.2       \\
\multirow{2}{*}{Web-mined}   & \textbf{WikiLingua}       & Guides          & 17,904        & Multi      & 407.4      & 24.6      & Multi      & 50.3       & 5.2       \\
                             & \textbf{CrossSum}         & News            & 4,975         & Multi      & 673.5      & 33.7      & Multi      & 44.8       & 1.21      \\
                             \cdashline{1-10}[.4pt/1pt]\noalign{\vskip 1pt}
Human-written                & \textbf{CroCoSum}         & News            & 18,557        & En         & 1079.6     & 55.4      & Zh         & 225.6      & 5.98      \\ \bottomrule
\end{tabular}%
}
\caption{Data statistics of CroCoSum and other CLS datasets. Except for En2ZhSum and CroCoSum, statistics are calculated based on the English-Chinese subsets.}
\label{tab:cls_stats}
\end{table*}

%% file: csmetrics_stats.tex
% \usepackage{booktabs}
% \usepackage{multirow}
% \usepackage{graphicx}

\begin{table*}[!ht]
\centering
\resizebox{\linewidth}{!}{%
\begin{threeparttable}
\begin{tabular}{@{}rrrrrrrrrr@{}}
\toprule
 &
   &
   &
   &
  \multicolumn{2}{c}{\textbf{Switched}} &
  \multicolumn{2}{c}{\textbf{CMI}} &
  \multicolumn{2}{c}{\textbf{SP}} \\ \cmidrule(l){5-6} \cmidrule(l){7-8} \cmidrule(l){9-10}  
\textbf{Task} &
  \textbf{Dataset} &
  \textbf{Total Sents} &
  \textbf{Avg Sent Len} &
  Sents &
  \% &
  All &
  Switched &
  All &
  Switched \\ \midrule
\multicolumn{10}{c}{\textbf{Cross-Lingual}}                                                            \\ \midrule
\multirow{5}{*}{Summ.} &
  \textbf{En2ZhSum} &
  1,336,155 &
  23.41 &
  90,930 &
  6.81 &
  0.49 &
  2.26 &
  0.14 &
  0.65 \\
              & \textbf{XSAMSum}   & 31,517  & 24.41 & 1,429  & 4.53  & 0.39  & 4.29  & 0.10 & 1.12 \\
              & \textbf{XMediaSum40k}     & 45,966  & 24.40 & 3,140  & 6.83  & 0.38  & 4.27  & 0.14 & 1.60 \\
              & \textbf{WikiLingua}      & 92,433  & 9.75  & 6,327  & 6.84  & 1.19  & 5.53 & 0.13 & 0.61 \\
              & \textbf{CrossSum}   & 6,027   & 36.96 & 2,099  & 34.83 & 2.10  & 4.97  & 0.94 & 2.23 \\ \midrule
\multicolumn{10}{c}{\textbf{Code-Switching}}                                                           \\ \midrule
Summ. & \textbf{GupShup}\tnote{$\alpha$}       & 76,330  & 10.07 & 43,407 & 56.87 & -     & -     & -    & -    \\
Tweet LID     & \textbf{EMNLP2014}\tnote{$\alpha$}    & 999     & 17.45 & 322    & 32.23 & 4.19  & 13.01 & 0.7  & 2.18 \\
Speech Recog.  & \textbf{ASCEND}\tnote{$\beta$}      & 12,314  & 11.83 & 3,326  & 27.01 & 5.02  & 18.59 & 0.62 & 2.28 \\
Speech Recog.  & \textbf{SEAME}  \tnote{$\beta$$\gamma$}     & 11,852  & 12.69 & 6,468  & 54.57 & 14.11 & 25.86 & 1.84 & 3.37 \\ \midrule
Summ. & \textbf{CroCoSum}     & 110,534 & 37.88 & 61,678 & 55.75 & 4.74  & 5.09  & 2.18 & 2.35 \\ \bottomrule
\end{tabular}%
\begin{tablenotes}
\item[$\alpha$] Due to limited access to data, we report statistics of GupShup based on its original paper \citep{mehnaz2021gupshup} and EMNLP14 based on \citet{gamback-das-2016-comparing}. "-" means statistics cannot be computed from the original paper. 
\item[$\beta$] We use transcriptions of the speech utterances, and remove noise tokens like <v\_noise> and [UNK] prior to calculation.
\item[$\gamma$] The training split of SEAME is non-public, the statistics are reported on its dev splits.
\end{tablenotes}
\end{threeparttable}
}
\caption{Code-switching metrics of CroCoSum and other CLS datasets.}
\label{tab:csmetrics_stats}
\end{table*}

%% file: baseline.tex
% \usepackage{booktabs}
% \usepackage{graphicx}
\begin{table*}[ht]
\centering
\resizebox{\linewidth}{!}{%
\begin{threeparttable}
\begin{tabular}{@{}rrrrrrrrrrrr@{}}
\toprule
 &
  \textbf{R1} &
  \textbf{R2} &
  \textbf{RL} &
  \textbf{BS} &
  \textbf{Sents.} &
  \textbf{Words} &
  \textbf{$\text{Sents}_{cs}$ \%} &
  \textbf{$\text{CMI}_{all}$} &
  \textbf{$\text{CMI}_{cs}$} &
  \textbf{$\text{SP}_{all}$} &
  \textbf{$\text{SP}_{cs}$} \\ \midrule
\textbf{GT}\tnote{$\alpha$}      & -     & -     & -     & -     & 5.96 & 225.05 & 55.28          & 4.75           & 5.11           & 2.19           & 2.36  \\ \midrule
\multicolumn{12}{c}{\textbf{Pipeline - Trans-Sum}}                                                                                           \\ \midrule
\textbf{mT5}     & 26.44 & 11.80 & 24.33 & 51.10 & 4.34 & 158.15 & -1.95          & +0.06 & +1.01 & \textbf{+0.08}         & +0.53 \\
\textbf{mBART}   & 34.97 & 16.39 & 30.72 & 57.70 & 5.80 & 240.89 & +2.36          & -0.10 & -0.13 & +0.19          & +0.19 \\
\textbf{mBART50} &
  34.91 &
  16.77 &
  30.85 &
  57.77 &
  5.67 &
  228.29 &
  +2.62 &
  \textbf{+0.03} &
  \textbf{+0.01} &
  +0.20 &
  +0.20 \\ \midrule
\multicolumn{12}{c}{\textbf{Pipeline - Sum-Trans}}                                                                                           \\ \midrule
\textbf{mT5}     & 24.93 & 11.43 & 23.08 & 47.08 & 7.03 & 174.98 & -15.81         & -0.67          & +0.11          & -0.99          & -0.83 \\
\textbf{mBART}   & 35.08 & 16.52 & 30.82 & 54.90 & 7.64 & 233.26 & -7.12          & -0.23 & -0.24          & -0.51          & -0.54 \\
\textbf{mBART50} & 35.27 & 17.01 & 31.26 & 54.94 & 6.97 & 195.57 & -8.08          & -0.27          & -0.26          & -0.63          & -0.67 \\ \midrule
\multicolumn{12}{c}{\textbf{End-to-End}}                                                                                                     \\ \midrule
\textbf{mT5}     & 31.62 & 15.87 & 28.85 & 53.79 & 4.56 & 165.27 & \textbf{+0.37} & +0.74          & +1.28          & +0.28          & +0.52 \\
\textbf{mBART}   & 38.44 & 19.94 & 34.01 & 58.67 & 5.33 & 193.77 & +4.89          & +0.86          & +0.81          & +0.18 & \textbf{+0.14} \\
\textbf{mBART50} &
  \textbf{38.73} &
  \textbf{20.34} &
  \textbf{34.35} &
  \textbf{58.81} &
  5.22 &
  189.08 &
  +7.41 &
  +1.31 &
  +1.26 &
  +0.29 &
  +0.25 \\ \midrule
\multicolumn{12}{c}{\textbf{Zero-Shot}}                                                                                                      \\ \midrule
\textbf{GPT-3}   & 26.50 & 12.33 & 24.00 & 52.08 & 4.74 & 153.53 & -16.37         & -1.62          & -0.75 & -0.95          & -0.62 \\ 
\textbf{\text{GPT-3.5}}   & 34.98 & 17.75 & 31.17 & 55.53 & 8.42 & 284.70 & -10.76  & -1.08         & -0.80 & -0.72          & -0.67 \\\bottomrule
\end{tabular}%
\begin{tablenotes}
\item[$\alpha$] GT refers to the ground truth summaries in the test set. ROUGE and BERTScore are omitted since there's no prediction to compare to.
\end{tablenotes}
\end{threeparttable}
}
\caption{Experimental results of different CLS baseline approaches.}
\label{tab:baseline}
\end{table*}

%% file: augmentation_results.tex
\begin{table}[t]
\centering
\resizebox{\columnwidth}{!}{%
\begin{tabular}{@{}rrrr@{}}
\toprule
\textbf{Method}                          & \textbf{R1} & \textbf{R2} & \textbf{RL} \\ \midrule
$\textbf{FT}_{\text{CroCo}}$                        & 38.73       & 20.34       & 34.35       \\
$\textbf{PT}_{\text{Wiki}}$ + $\textbf{FT}_{\text{CroCo}}$                       & 37.25       & 19.09       & 33.05       \\
$\textbf{PT}_{\text{Cross}}$ + $\textbf{FT}_{\text{CroCo}}$                         & 37.85       & 19.36       & 33.47       \\
$\textbf{PT}_{\text{Wiki + Cross}}$ + $\textbf{FT}_{\text{CroCo}}$            & 37.19       & 18.81       & 32.79       \\
$\textbf{FT}_{\text{Wiki + Cross + CroCo}}$ & 37.91       & 19.58       & 33.82       \\ \bottomrule
\end{tabular}%
}
\caption{Result Comparison of No Data Augmentation vs.\ Additional CLS Pretraining. \textbf{PT} means pretrain and \textbf{FT} means finetune.}
\label{tab:augmentation}
\end{table}

%% file: appendix.tex
\clearpage
\onecolumn
\appendix
\section{Data Examples}
\label{sec:examples}
Here are two data examples from the train set of the dataset. Due to the length of the source articles, the texts in the \texttt{link\_body} field are truncated to fit within the page.

\paragraph{Single-Source Example}
\begin{verbatim}
 {'post_id': '70292',
'post_url': 'anon_post_link_1',
'post_title': '亚马逊准备制作辐射电视剧',
'post_body': '亚马逊 Amazon Prime Video 准备制作改编自经典游戏《辐射》系列的电视剧。《辐射》系列设定发生在 22 世纪，其背景受到了 1950 年代核恐慌的影响，2077 年中美之间因争夺能源而发生了大规模核战，世界遭到毁灭，故事就发生在核战一百年后，藏身于避难所的人走出地下探索核战后的世界。亚马逊在 2020 年获得了 《辐射》的电视改编权，HBO 《西部世界》系列的创作人Lisa Joy 和 Jonah Nolan 夫妇担任监制，剧本由 Geneva Robertson-Dworet 和 Graham Wagner 负责。Robertson-Dworet 之前的作品包括 《惊奇队长》和《古墓丽影》，以及将在 2023 年上映的《星际迷航》新作。Wagner 的作品主要为喜剧片《Portlandia 》、《The Office》、《Silicon Valley》和《Baskets》。《辐射》电视剧的叙述风格将融合 Robertson-Dworet 的科幻动作和 Wagner 的喜剧。',
'links': {'link_id': ['49547'],
'link_url': ['anon_source_link_1'],
'link_title': ['Amazon’s Fallout TV series is about to enter production'],
'link_body': [
"Amazon Prime Video's adaptation of the Fallout franchise of video games is 
entering production this year, and its two lead writers have been named, 
according to reports in Deadline and Variety....Though the TV series begins 
production this year, we don't yet know when it will start streaming 
to audiences."]}

\end{verbatim}

\paragraph{Multi-Source Example}
\begin{verbatim}
{'post_id': '57688',
'post_url': 'anon_post_link_2',
'post_title': 'N.K. Jemisin 连续第三年摘下雨果最佳长篇',
'post_body': '黑人女作家 N. K. Jemisin 的《破碎星球》三部曲之大结局《巨石苍穹》连续第三年摘下雨果最佳长篇。该系列的前两部《第五季》和《方尖碑之门》分别赢得了 2016 年和 2017 年的雨果奖。这三部小说的中文版已经出版，已经售出电视剧改编权。此前《巨石苍穹》还赢得了星云奖，她跻身于极少数连续赢得雨果奖以及同时获得雨果星云双奖的科幻作家行列。2018 年度的雨果奖获奖名单包括：最佳长篇小说：N.K. Jemisin 的《巨石苍穹》；最佳中长篇小说：Martha Wells 的《All Systems Red》；最佳中短篇小说：Suzanne Palmer 的《The Secret Life of Bots》；最佳短篇小说：Rebecca Roanhorse 的《Welcome to your Authentic Indian Experience™》：新设的奖项最佳系列小说：Lois McMaster Bujold 的《World of the Five Gods》；最佳幻想电视：《善地(The Good Place)》之《The Trolley Problem》 ；最佳科幻电影：《神奇女侠》。',
'links': {'link_id': ['28857', '28858', '28860', '28861', '28862', '28863'],
'link_url':['anon_source_link_2','anon_source_link_3','anon_source_link_4',
'anon_source_link_5','anon_source_link_6','anon_source_link_7'],
'link_title': ['The Stone Sky',
'The Fifth Season (novel)',
'2018 Hugo Award Winners', 'All Systems Red',
'Clarkesworld Magazine - Science Fiction & Fantasy',
'Welcome to Your Authentic Indian Experience™'],
'link_body': [
'The Stone Sky is a 2017 science fantasy novel by American writer N. K. 
Jemisin. It was awarded the Hugo Award for Best Novel,[2][3] the Nebula 
Award for Best Novel, [4]and the Locus Award for Best Fantasy Novel[5] in 
2018...Finishing The Stone Sky left me utterly breathless by the scale 
and scope of what Jemisin accomplished in these three books—narratively, 
technically, and thematically.[6]', 

'The Fifth Season is a science fiction fantasy novel by N. K. Jemisin.[1]
[2]The book was released on August 4, 2015 by Orbit Books....This is the 
first part of The Broken Earth Trilogy.[6] The second novel in the trilogy, 
The Obelisk Gate, was published on August 16, 2016.',

'Here are the results of the 2018 Hugo awards. Congratulations to all of 
the of the winners....Sarah Kuhn* Jeannette Ng Vina Jie-Min Prasad 
Rivers Solomon',

'All Systems Red is a 2017 science fiction novella by American author 
Martha Wells. The first in a series called The Murderbot Diaries,...Mensah 
later mentions that GoodNightLander—the contractor of Murderbot's Milu 
clients—also wants to hire "Security Consultant Rin". Murderbot is content 
to know that it has options.',

'I have been activated, therefore I have a purpose, the bot thought...“But 
we are home,” 4340 said, and Bot 9 considered that that was, any way you 
calculated it, the truth of it all.',

'In the Great American Indian novel, when it is finally written, all of the 
white people will be Indians and all of the Indians will be ghosts....her 
article Decolonizing Science Fiction and Imagining Futures: An Indigenous 
Futurisms Roundtable can be found in Strange Horizons... ']}
\end{verbatim}

\section{Baseline Details}
\label{sec:exp_details}
In the pipeline and end-to-end methods, we pick the best-performing checkpoint based on the validation set for inference after finetuning multilingual models for 15 epochs with a learning rate of 5e-05 and a batch size of 4. For the zero-shot method, we used text-davinci-002 for GPT-3 and gpt-3.5-turbo for GPT-3.5 with a temperature of 0.7 and top\_p of 1.